\documentclass[conference]{IEEEtran}
\IEEEoverridecommandlockouts
\usepackage{cite}
\usepackage{amsmath,amssymb,amsfonts,array, url}
\usepackage{caption}
\usepackage{subcaption}
\usepackage{algorithmic}
\usepackage{graphicx}
\usepackage{textcomp}
\usepackage{xcolor}
\usepackage{orcidlink}
\def\BibTeX{{\rm B\kern-.05em{\sc i\kern-.025em b}\kern-.08em
    T\kern-.1667em\lower.7ex\hbox{E}\kern-.125emX}}
\begin{document}

\title{TONUS: Neuromorphic human pose estimation for artistic sound co-creation\\
\thanks{This project has been funded funded by the Bayerische Forschungsstiftung under the grant number AZ-1558-22}
}

\author{\IEEEauthorblockN{1\textsuperscript{st} Jules Lecomte}
\IEEEauthorblockA{\textit{Neuromorphic Computing} \\
\textit{Fortiss GmbH}\\
Munich, Germany \\
\orcidlink{0000-0002-7103-0843}}
\and
\IEEEauthorblockN{2\textsuperscript{nd} Konrad Zinner}
\IEEEauthorblockA{\textit{Music, Komposition Film und Medien} \\
\textit{Hochschule für Musik und Theater }\\
Munich, Germany \\
konrad.zinner@web.de}
\and
\IEEEauthorblockN{3\textsuperscript{nd} Michael Neumeier}
\IEEEauthorblockA{\textit{Neuromorphic Computing} \\
\textit{Fortiss GmbH}\\
Munich, Germany \\
\orcidlink{0000-0002-9601-7477}}
\and
\IEEEauthorblockN{4\textsuperscript{rd} Axel von Arnim}
\IEEEauthorblockA{\textit{Neuromorphic Computing} \\
\textit{Fortiss GmbH}\\
Munich, Germany \\
\orcidlink{0000-0002-9680-7591}}}

%\author{\IEEEauthorblockN{Anonymous Authors}}
\maketitle

\begin{abstract}
Human machine interaction is a huge source of inspiration in today's media art and digital design, as machines and humans merge together more and more. Its place in art reflects its growing applications in industry, such as robotics.
However, those interactions often remains too technical and machine-driven for people to really engage into. On the artistic side, new technologies are often not explored in their full potential and lag a bit behind, so that state-of-the-art research does not make its way up to museums and exhibitions.
Machines should support people's imagination and poetry in a seamless interface to their body or soul.
We propose an artistic sound installation featuring neuromorphic body sensing to support a direct yet non intrusive interaction with the visitor with the purpose of creating sound scapes together with the machine.
We design a neuromorphic multihead human pose estimation neural sensor that shapes sound scapes and visual output with fine body movement control. In particular, the feature extractor is a spiking neural network tailored for a dedicated neuromorphic chip.
The visitor, immersed in a sound atmosphere and a neurally processed representation of themselves that they control, experience the dialogue with a machine that thinks neurally, similarly to them.
\end{abstract}

\section{Introduction}
Neuromorphic computing is a brain-inspired approach to computing, sensing and hardware. It can be applied to the machine learning realm with Spiking Neural Networks (SNN), the so-called third generation of neural networks which rely on time dependent, biologically inspired neuron dynamics and represent information with sparse and asynchronous spikes. Neuromorphic vision sensors, or event based cameras are novel sensors that detect per pixel brightness changes asynchronously \cite{dvs}, similarly to the human eye. Such properties enable to capture motion at high spatio-temporal resolution while remaining data and power efficient. Event based cameras provide events that can be fed to a SNN without encoding overhead. Additionally, SNNs can be run on neuromorphic hardware in a sparse and energy efficient fashion. Put together, a event based camera and a SNN running on a dedicated chip makes a very promising candidate for visual edge computing.
Human Pose Estimation (HPE) consists of localizing human body poses, and body joints (head, shoulders, legs...) and gives a skeleton representation of the body, often from a camera input. HPE can be used in numerous applications: biomechanical analysis, autonomous driving or human machine interaction. In the latter, a user can intuitively interact with a large span of actions: not only can specific gestures be mapped to specific commands by the machine, but joints' positions and amplitudes are also a means to modulate and quantize desired actions. 

SNNs have already been used in common machine learning tasks, such as object detection, optical flow or image classification. However they are hard to train due to their inherent non-derivatibility and the need to account for the time dimension \cite{shrestha2018slayerspikelayererror}. Thus, no fully spiking networks for HPE have been ported on a neuromorphic device yet. This paper presents a human machine sound design art installation integrating neuromorphic components (an event based camera, a SNN and a neuromorphic processor) to perform HPE. %for musical creation.

The scientific goal of this work is to design a real time sensor for human pose estimation taking events from a neuromorphic vision sensor as inputs. A neural network predicts the pose and uses a SNN as an encoder, designed to be run on Intel Loihi 2 \cite{loihi2}. The overall model is small, energy efficient and accurate enough to be use in an interactive artistic context.

The artistic goal is a scientific-artistic sound installation that connects people and machine through sound and lights. It explores the dialogue between man and neuromorphic machines, which present human-like processing properties. The perception of the visitor by the machine, and of the machine by the visitor are both based on attention and neural communication. Both influence the soundscape in an audiovisual feedback loop and this exchange creates unique atmospheres. The visual output is created from the artificial brain activity and blurs the technology into sensations.

To the best of our knowledge, this work is the first human pose estimation network able to run partly on a neuromorphic chip. It also outperforms state of the art networks in terms of floating operations. Artistically, it is the first prototype of a human machine interaction based on neuromorphic sensing.

\section{Related Work}

 \subsection{Pose estimation}

HPE has seen a growing interest with the advent of deep learning and convolution networks: such techniques enable accurate single or multiple persons localization from visual input, not only of the full body but also limbs.
Top down models first localize humans in the field of view and try to detect body joints in the active area \cite{alphapose}. Bottom-up methods directly recognize joints and, for multi-person tracking, associate joints together in a second stage, using for instance bipartite matching algorithms \cite{multiperson2d}. Another work relies on a multihead architecture to provide joint and body center heatmaps as well as joint regression from the body center \cite{zhou2019objectspoints}. This approach implicitly associates joints to one person in case of a multi-person use case and refines the localization as joints are localized with regression and heatmaps simultaneously. Google's MoveNet \cite{movenet} builds upon it and proposes a small footprint network for edge devices. Motivated by the power efficiency and the high temporal resolution of event based cameras, the DHP19 dataset provides stereo recordings of both RGB and event based cameras \cite{Calabrese_2019_CVPR_Workshops} for action recognition and HPE with joints annotated by markers. An Artificial Neural Network (ANN, non spiking) is also given as a reference implementation for both 2D and 3D HPE. Other ANNs have been proposed for event based HPE \cite{Goyal_2023_CVPR}\cite{TORE}, underlying the importance of a relevant event representation to preserve relevant information while discarding noise. %\textbf{unpublished work:} 3D HPE Spiking transformer \cite{zou2021eventhpeeventbased3dhuman} which also introduces its dataset SynEventHPD and train other SNN models on MMHPSD as a benchmark.
Another work focuses on 3D human pose estimation with a dance application and its own event based dataset \cite{danceHPE}. \cite{aydin2024hybridannsnnarchitecturelowpower} describes a hybrid ANN-SNN network  where an ANN is used at a lower frequency to initalize SNN neurons' states of a similar architecture. This implementation highlights its fast and low energy HPE solution but does not consider constraints arising when porting networks on neuromorphic hardware.

 \subsection{ Visual human-machine interaction}
With increasing performances of human pose estimation solutions, many applications in human machine interactions have been envisioned. In robotics, some works focus on having robots replicate what they see \cite{3dposeforrobotic}, and others handle situations with obstructions or intersections of body parts, to enable realistic service oriented robotics \cite{bodytrackerrealistic}. Industrial applications emerge, with robotic arms relying on finger ray tracing and safe space detection in soldering scenarios \cite{maira}. Hand pose recognition is the focus of many works, since they already play a key role in communicating human intentions, for instance with sign language \cite{signlanguage}. On a lower level, event based cameras have already been used to estimate hand pose in \cite{rudnev2021eventhands}.
The release of the Microsoft Kinect in 2010 opened the door to many human pose based games such as Kinect Adventures \cite{kinectadventure} and Kinect fun lab: Air band \cite{kinectairband}.

\subsection{Pose and gesture driven sound installations}
\subsubsection{Closest sound installation concepts}
Using body pose and hand gestures to generate or alter sound has been a classic engineering, research and artistic topic, especially since the Kinect was released. It has been an obvious application for gaming, as mentioned before, but artistic or "gamo-artistic" installations have surfaced quite soon too, like Kling Klang Klong's Trigger Playground Experiment \cite{kkktrigger}. This installation is very similar to ours, but with a commercial image-based data- and power-hungry sensor, and a basic technical interface without an artistic standpoint. It is closer to gaming than to arts. Further works by Kling Klang Klong have explored much deeper artistic atmospheres of body-driven sound installations, like 2015's Momentum \cite{kkkmomentum}, where visual effects generation seems though to overrule sound generation in attracting the visitor's attention. Or 2018's Disco Dusche \cite{kkkdisco}, which is much closer to our concept, without the visual neural feedback concept.
\subsubsection{Gesture-only interaction}
Other, yet more distant gesture-based interactive art can be mentioned like Karlheinz Essl's 2015 KlanDerWisch \cite{keklang}, where gestures directly produce sound in a very minimal set up without any visual feedback and limited artistic flair. Many others have been experimented, that also stay on the technical achievement level. Although a bit off topic, since it is not about sound generation, we would like to cite the highly dynamic and very immersive Box robotic installation from Bot\&Dolly \cite{bdbox} in 2013, where gesture and artistic feedback are taken to the technical apogee level.
\subsubsection{Sculptural approaches}
Real artistic statements have been proposed in the sculptural domain, with dynamic or static counterparts that interact tonally and/or visually with the visitor. A perfect example is Maria Koshenkova's and Richard Deutsch's 2013 glass sound installation \cite{rdglass}, where sculptural beauty merges with an interactive sound scape and takes the visitor inside the glass structure without anything moving. Emma Sharpe's 2016 Grasping Sound \cite{esgrasping} adds passive dynamics to her beautiful simple chain installation, while Riccardo Castagnola in 2016 reduces the sculptural object to a super minimal light point in his Mirror Of Truth \cite{rcmirror}.
All these installations are more museal than ours. Indeed, they are more static in the visitor expected behavior, as well as in the counterpart object's response.
\subsubsection{Interactive dancing}
Of course, on the other side of dynamics, interactive dancing has been explored since the inception of motion capture technologies. There are a lot of projects, Group Gravity's 2008 Noise and Silence \cite{ggnoise} being an early one. Interactive dancing is a close but different concept as ours, because it does not focus on science and technology as an artistic source, but uses tech devices as a tool for the dancing experience, which is the artistic focal point.

\section{Method}

\subsection{Sparse event-based pose estimation} \label{subsection:sparse event based hpe}
First, we describe how events are collected from the camera and processed in a spiking neural network to output pose estimation. We accumulate events in so-called time bins. Events are binned within a 10 milliseconds time window in a various number of bins of the same duration. For instance, it can be 8 bins of 1.25~ms or 16 bins of 625~ms. For each bin, the polarity (sign of the pixel brightness change) is kept in 2 channels but events are collapsed in time. Two representations are implemented, graded and binary spikes. Graded spikes take the number of events per polarity and coordinates within one time bin as values, while binary spikes take a non null value only if at least one event occurred for this polarity and coordinate in the time bin. \newline
The neuron model used for this network is the CUrrent BAsed leaky integrate and fire (CUBA) neuron for Loihi 2 \cite{loihi2}: 
    \begin{equation}
        u[t] = (1 - \alpha_u)\,u[t-1] + x[t]
    \end{equation}
    \begin{equation}
        v[t] = (1 - \alpha_v)\,v[t-1] + u[t]
    \end{equation}
    \begin{equation}
        s[t] = v[t] \geq \vartheta
    \end{equation}
    \begin{equation}
        h[t] = v[t]\,(1-s[t])
    \end{equation}
    with, at time-step $t$, $u$ the neuron current, $x$ the input, $s$ the output spikes, $v$ the membrane potential, $h$ the membrane potential after the trigger of a spike, $\vartheta$ the membrane threshold, $\alpha_u$ and $\alpha_v$ the current and the voltage decay respectively. \newline
Another implemented neuron is the Parametric Leaky Integrated and Fire (PLIF) from the \textit{spikingjelly} library \cite{spikingjelly}, with the following dynamics:
    \begin{equation}
        h[t] = f(v[t-1],  x[t])
    \end{equation}
    \begin{equation}
        s[t] = \Theta(h[t]-\vartheta)
    \end{equation}
    \begin{equation}
        v[t] = h[t](1-s[t])+v_{reset}s[t]
    \end{equation}
    
    wiht $\Theta$ the Heaviside step function and $v_{reset}$ the membrane potential value taken after emitting a spike.

\subsection{Network architecture} \label{subsection:architecture}

\begin{figure}
     \centering
     \begin{subfigure}[b]{0.24\textwidth}
         \centering
         \includegraphics[width=\textwidth]{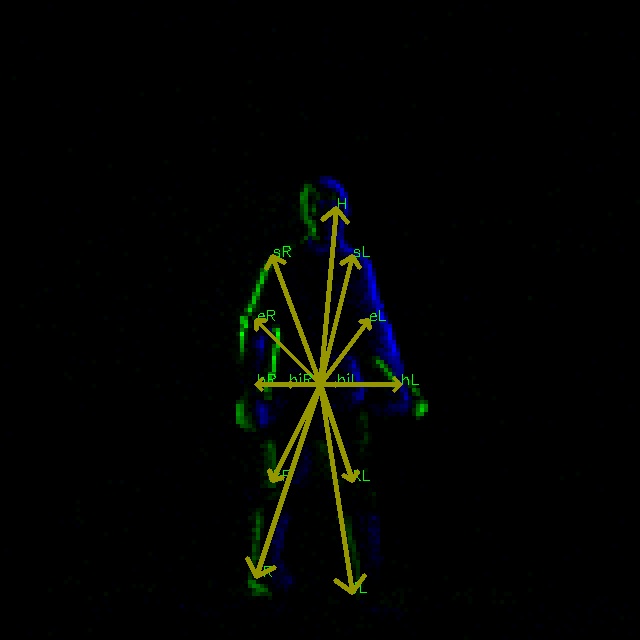}
         \subcaption{Joints regression (40,40,26)}
     \end{subfigure}
     \hfill
     \begin{subfigure}[b]{0.24\textwidth}
         \centering
         \includegraphics[width=\textwidth]{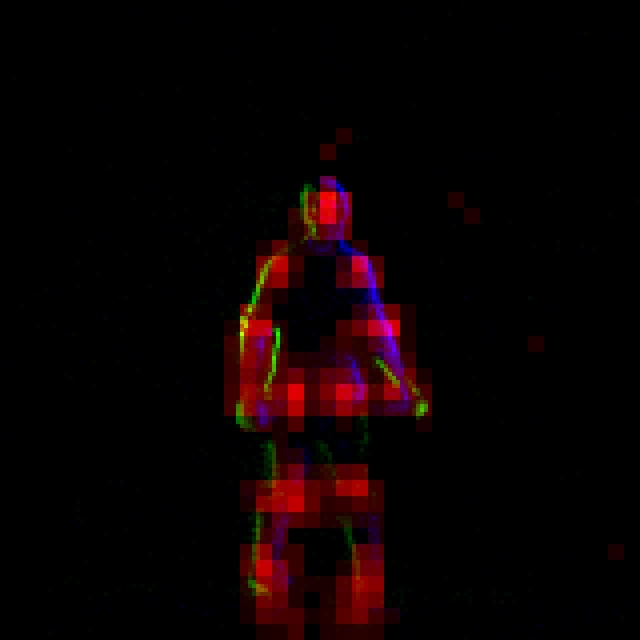}
         \subcaption{Joints heatmap (40,40,13)}
     \end{subfigure}
     \begin{subfigure}[b]{0.24\textwidth}
         \centering
         \includegraphics[width=\textwidth]{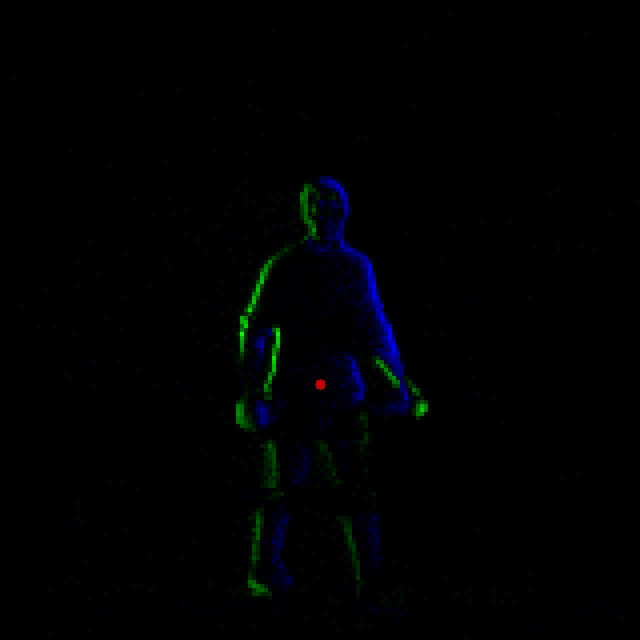}
         \subcaption{Center of mass (40,40,1)}
     \end{subfigure}
     \hfill
     \begin{subfigure}[b]{0.24\textwidth}
         \centering
         \includegraphics[width=\textwidth]{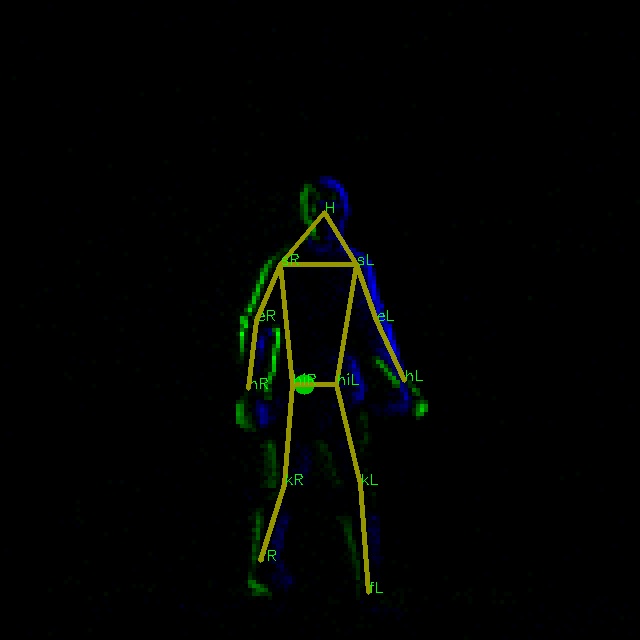}
         \subcaption{Ground truth positions}
     \end{subfigure}
        \caption{Visualisation of multihead outputs and ground truth overlayed on input events, with output map formats.}
        \label{fig:output format}
\end{figure}

A compact encoder-decoder architecture similar to \cite{Goyal_2023_CVPR} is used with a SNN encoder block and a ANN decoder block. At the end of the encoder, the membrane potential of the last convolution layer is fed as a floating value input to the decoder. The model is kept as small as possible to run in real time, as described in Tab.~\ref{tab:encoder-architecture}. ResN are ResNet blocks which consist of a convolution of dilation 2 and kernel size (3,3) and a BatchNorm2D layer repeated two times, DConv blocks are separable convolutions of kernel size 3 keeping a identical number of channel followed by convolutions of kernel size 1 with a different output channel. DConv convolutions are followed by a BatchNorm2d layer and a ReLU is applied at the end of the block. The overall architecture is depicted in Fig.~\ref{fig:architecture}. As in \cite{zhou2019objectspoints} and \cite{movenet}, a multihead architecture is chosen at the output, with a varying number of channels which representations are described in Fig.~\ref{fig:output format} and have the following structure:

\begin{figure}
     \centering
     \includegraphics[width=0.48\textwidth]{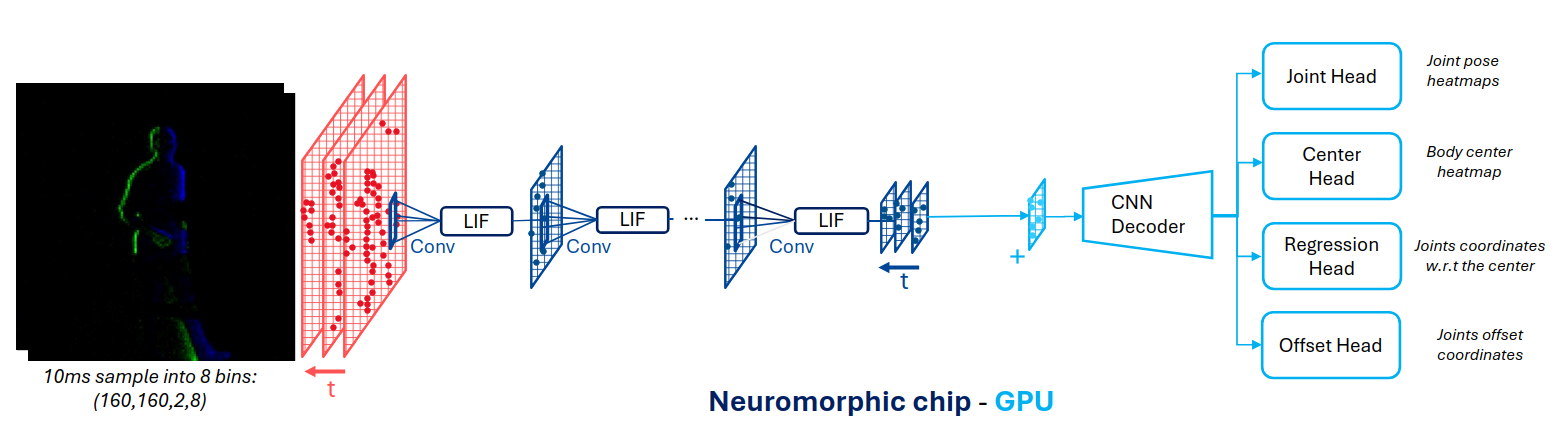}
    \caption{The model architecture with a spiking encoder.}
    \label{fig:architecture}
\end{figure}

\begin{itemize}
    \item \textbf{Heatmap head:} The model predicts 13 keypoints, and there is one heatmap channel for each keypoint, encoding the keypoint position probability. The ground truth is therefore a frame with a gaussian distribution centered around the downsampled keypoint ground truth position $(P_x//4,P_y//4)$.
    \item \textbf{Center head:} The center channel structure is similar to a heatmap channel and encodes the keypoints' barycenter $(C_x,C_y)$.
    \item \textbf{Regression head:} A regression channel is allocated for every keypoint coordinates axis. The ground truth is a zero valued frame except at the location $(C_x,C_y)$ with the keypoint distance to the barycenter $R_i$ on one axis, such that $P_i = (C_i + R_i) \times 4 + O_i$, where \textit{i} is in ${x,y}$.
    \item \textbf{Offset head:} As we decreased the input resolution by a factor 4, the offset head allows a sub pixel precision at the output, to adjust regression's results to the input resolution. It has 26 channels as well, each of them having at coordinate $(C_x,C_y)$ the offset value $O_i$, where i is in ${x,y}$.
\end{itemize}

\begin{table}[htbp]
\caption{Architecture details}
\setlength{\tabcolsep}{2pt} 
\begin{subtable}[h]{0.45\textwidth}
\caption{Encoder implemented for neuromorphic hardware}
\centering
\begin{tabular}{|l|cccccccc|}
\hline
Layer & 1 & 2 & 3 & 4 &5 & 6&7&8\\ 
\hline
Kernel size & 5 & 5&5&5&3&5&3&3\\
\hline
Output channels &16&32&32&64&64&128&128&128\\
\hline
Output H,W & 80 & 40 & 20 & 20 & 20 &10 & 10 & 10 \\
\hline
\end{tabular}
\end{subtable}
\setlength{\tabcolsep}{2pt} 
\begin{subtable}[h]{0.45\textwidth}
\caption{Decoder running on a GPU, with the two last layers duplicated in the 4 heads and different output representations.}
\centering
\begin{tabular}{|l|ccccccc|}
\hline
Layer & 1 & 2 & 3 & 4 &5 & 6&7\\ 
\hline
Kernel size & 5 & 3 &5&3&3&3&1\\
\hline
Output channels &64&64&32&32&24&96,96,96,96&13,1,26,26\\
\hline
Output H,W & 10 & 20 & 40 & 40 & 40 &40 & 40 \\
\hline
Type & ConvT & ResN & ConvT & ResN & Conv & DConv & Conv\\
\hline
\end{tabular}
\end{subtable}
\label{tab:encoder-architecture}
\end{table}

\subsection{Training details}
The CUBA network is implemented using Intel's lava-dl library and the SLAYER algorithm with Back Propagation Through Time (BPTT) \cite{shrestha2018slayerspikelayererror}. The PLIF network is implemented with SpikingJelly and its BPTT framework \cite{spikingjelly}.
The training is done on a GPU NVIDIA 2080 Ti for 20 epochs. Input resolution is set to 160x160 to keep a fast and efficient implementation.
The loss is inspired from CenterNet \cite{zhou2019objectspoints} and is computed differently for each head: a focal loss for the heatmap head to account for the unbalanced ground truth, a mean square error for the regression and offset, and a weighted mean squared error for the center head.

For our installation, we trained the model with our own dataset, recorded with a Prophesee Camera IM636 Evk4. The ground truth was captured with the Simi markerless motion system \cite{simisystem} and consists of the 100~Hz 13 annotated body joints used in the DHP19 dataset \cite{Calabrese_2019_CVPR_Workshops}. The output resolution was 1280*640 with 12 subjects recorded. Unlike the DHP19 dataset, subjects move freely with diverse angles and distances to the camera, to prevent overfitting. The sensor is the same in the live installation to ensure a consistent data distribution and increase the accuracy.

\subsection{Hardware aware optimization}

We designed our encoder to be run on the Intel Loihi 2 \cite{loihi2} with its energy and latency benefits. However, running AC operations at a low energy cost constrains data and spikes digital representations. The weights are 32~bits quantized during training to cope with the chip resolution. Within the network, spikes must be binary except for the first layer which can take either graded or binary spikes as inputs, justifying our event representation choice described in Sec.~\ref{subsection:sparse event based hpe}. 
Considering that neuromorphic hardware is still in its infancy, some synaptic connections implementing classical deep learning blocks are not available yet. In particular, skipped connections and deconvolutions are not implemented, preventing us to fully reproduce the Movenet \cite{movenet} architecture on hardware.

\subsection{Kalman filtering}
Similarly as in \cite{Goyal_2023_CVPR}, a Kalman filter is used to ensure a more stable estimation of the joint position and addresses the invisible joint issue if no events are captured. A four dimensions Kalman state is used for each joint, with two components for the 2D position and  two for the the 2D velocity. The velocity is considered constant and the motion linear, such that its prediction is equal to the previous value and the predicted position is the previous value plus the velocity times the elapsed. For every new model prediction, joint coordinates are used as state measurement.

\section{Artistic installation}

\subsection{Artistic statement}

For the first time, motion tracking is performed using neuromorphic computing, an emerging AI technology inspired by the human brain, simulating the very details of biological neurons, in contrast to conventional AI. This makes the machine very similar to humans.
A neuromorphic event-based camera transmits the signatures of movements rather than entire images. The person is de facto unrecognizable, only the traces of their movements are.
The neural network of spiking neurons, runs on a neuromorphic chip, that mimics a biological brain. Our goal is to confuse the visitor by showing this similarity via sound and light feedback, by visually representing the activity of the machine's neurons in an abstract way and blurring the technological separation.
Another special feature of neuromorphic technology is its very low energy consumption. It is also referred to as "Green AI". The neuromorphic chip is presented in such a way that it takes center stage and, thanks to its small dimensions and low energy consumption, its staging will evoke the frugality of green artificial intelligence.
The projected visual feedback is the direct output of the camera (blue and green shape edges), plus the “heatmap” produced by a spiking CNN based neural network.
The LED installation reacts to the heatmap values of active joints, which also trigger sound clicks, enabling the visitor to hear and see neuronal spiking activity.
\subsection{Installation set-up}

\begin{figure}
     \centering
     \begin{subfigure}[b]{0.24\textwidth}
         \centering
         \includegraphics[width=\textwidth]{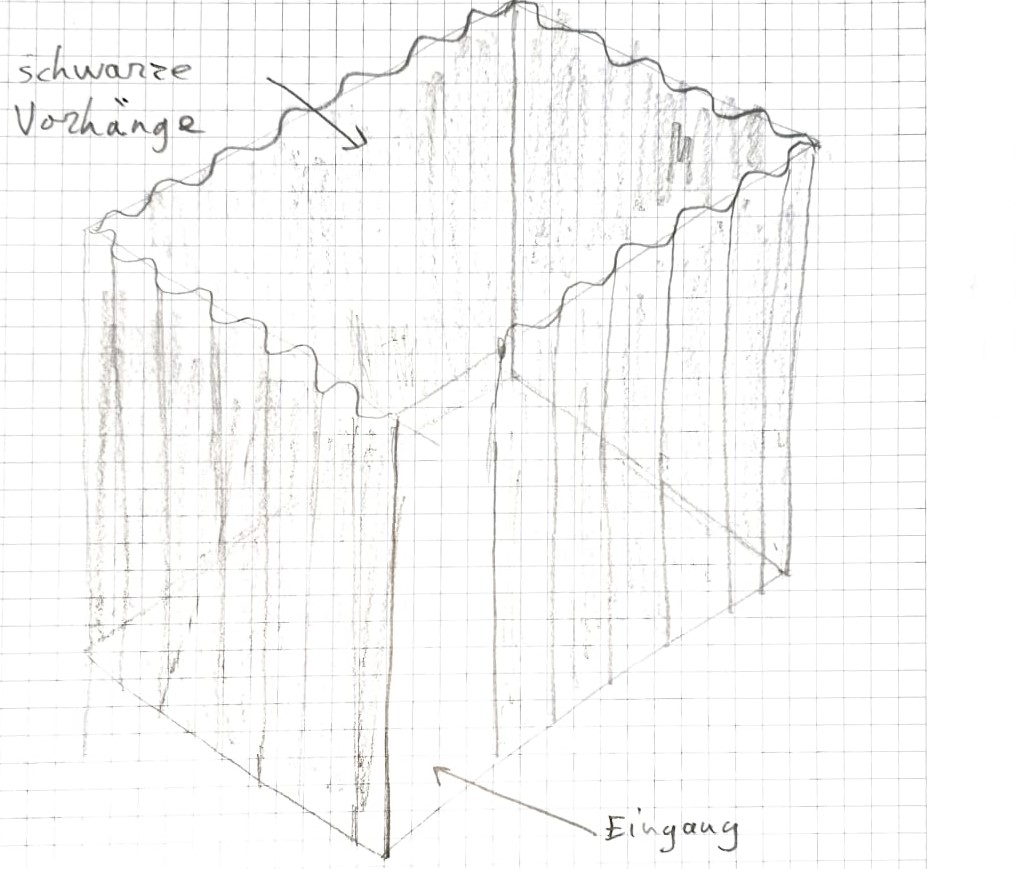}
         \subcaption{From the outside}
     \end{subfigure}
     \hfill
     \begin{subfigure}[b]{0.24\textwidth}
         \centering
         \includegraphics[width=\textwidth]{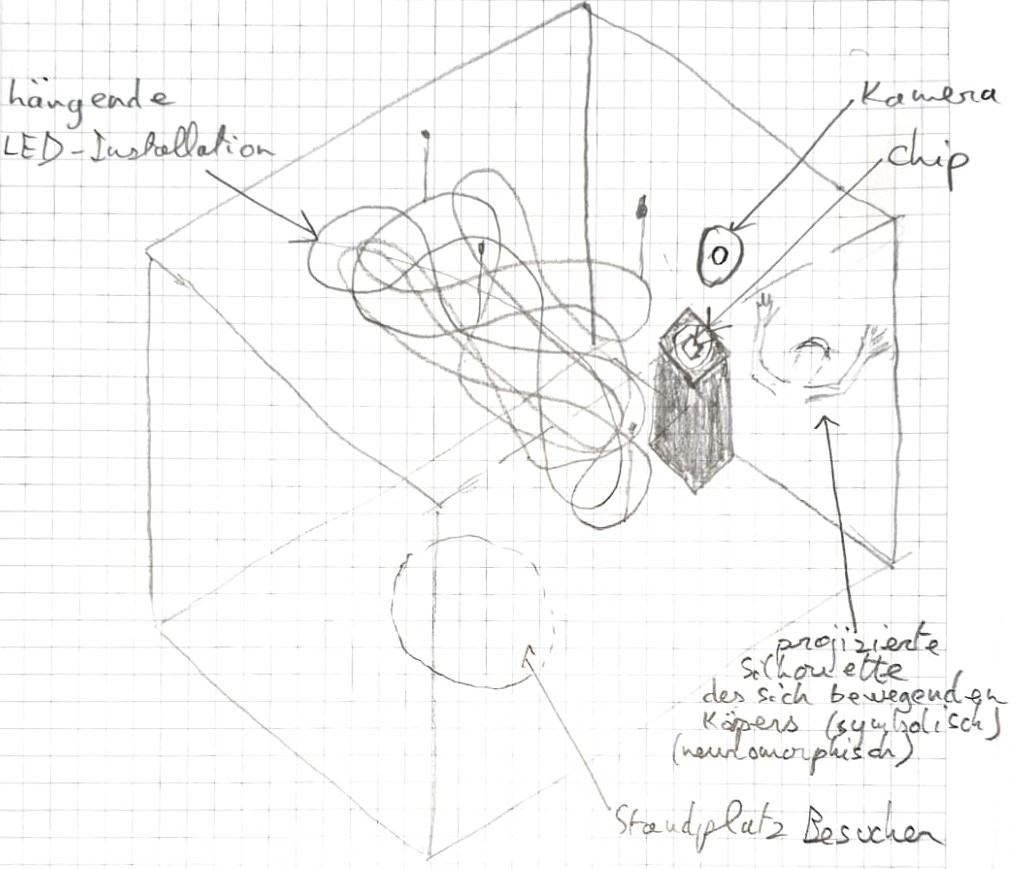}
         \subcaption{From the inside}
     \end{subfigure}
    \caption{The inside arrangement as it was in the original plan. The LED "cloud" moved to inside the totem, the camera moved to the totem desktop and the projected "silhouette" (heatmap) moved to the left curtain wall.}
    \label{fig:booth-schema}
\end{figure}

\begin{figure}
     \centering
     \begin{subfigure}[b]{0.24\textwidth}
         \centering
         \includegraphics[width=\textwidth]{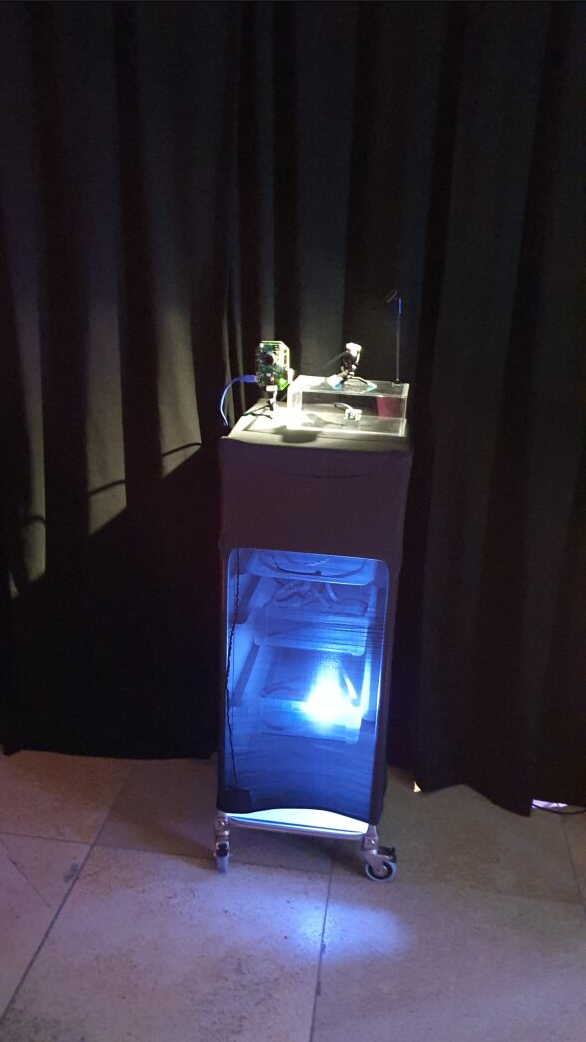}
     \end{subfigure}
     \hfill
     \begin{subfigure}[b]{0.24\textwidth}
         \centering
         \includegraphics[width=\textwidth]{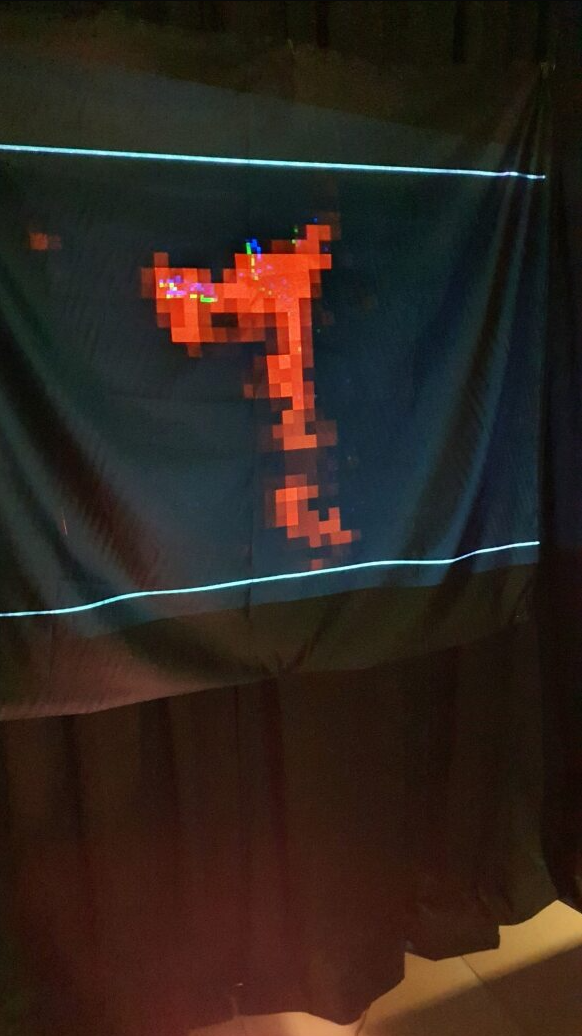}
     \end{subfigure}
        \caption{Left: the totem presenting the neuromorphic hardware in a jewel showcase and flashing upon neuronal activity. Right: a neuronal representation of the perceived body heatmap.}
        \label{fig:totem-screen}
\end{figure}

The installation is presented as a sound box protected by a black curtain, with enough space for visitors to move their whole body (2m x 3m). The dark and hushed atmosphere within the box gives the visitor intimacy and encourages free motion. Moreover, it ensures for the senses to stay focused on sound and the visual neuronal feedback of the art work. Indeed, as depicted in the Fig.~\ref{fig:booth-schema}, that we successfully submitted to the Festival der Zukunft in Munich, there is a neural totem facing the visitor, as well as a projected neuronal body representation of him. The totem not only carries the museally presented neuromorphic chip Loihi2 and the event camera, but also flashes with neuronal activity. A spool of LED ribbon is embedded into the totem and flashes through a translucent black spanned cover. Just like in the visitor's brain, different parts of the totem flash in real time when different visitor body parts are being perceived. This creates a visual neuronal dialogue between the visitor and the thinking machine. On the left curtain, the output heatmap (regions of activity in the visitor's body) is projected in red over black, as shown on Fig.~\ref{fig:totem-screen} and creates dynamics in the installation, while giving confidence to the visitor that their movements are being understood. This creates intimacy between the visitor and the machine.

\begin{figure}
    \centering
    \includegraphics[width=1\linewidth]{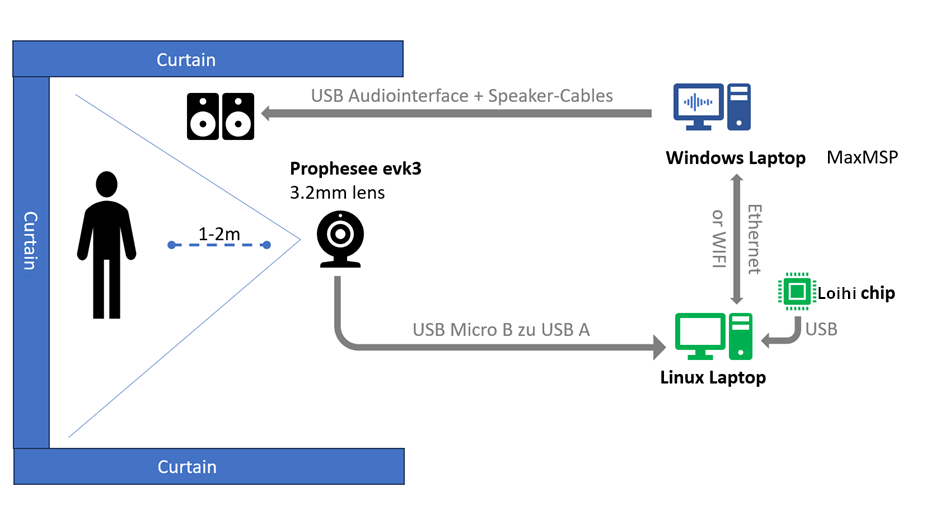}
    \caption{Technical setup of the installation.}
    \label{fig:tech_setup}
\end{figure}

For reference, the technical rider is provided in Tab.~\ref{table:tech_rider} and Fig.~\ref{fig:tech_setup} details the setup.

\begin{table}[htbp]
\centering
\caption{Description of the installation}
\begin{tabular}{|l| m{0.35\textwidth}|}
\hline
Component & Description\\ 
\hline
Sensing & Events are sensed by a Prophesee Evk3 camera with a wide angle 3.2mm focal length.\\
 &  The camera output is sent via USB to a linux computer preprocessing events.\\
 &  To decode the latent space from the neuromorphic sensor, a NVIDIA GPU 2080 Ti is used.\\
\hline
Audio & A laptop for sound design and musical logics with Max and Ableton software retrieves pose estimation values over a UDP socket.\\ %slurm-4840.out
 & A digital to analog mixer converts digital audio to the speakers.\\
\hline
Visual & A 10m white LED ribbon that can be addressed and controlled over DMX.\\
&An Arduino board, connected with the sound laptop gets heatmap values over UDP to control the LED ribbon: light is propagated around given LED positions mimicking the spike propagation and synchronized with sound clicks.\\ %slurm-4840.out
& A projector displays the heatmap predictions of visitors' joint inside the box.\\ %slurm-4768.out
& A monitor shows the visitor's skeleton outside of the installation.\\
\hline
\end{tabular}
\label{table:tech_rider}
\end{table}

\subsection{Musical landscapes}
The installation proposes different sound atmospheres. It localizes the visitor's skeleton with a neuromorphic chip and camera, in a energy efficient and sparse way. Visual and audio feedbacks from the body motion exposes the neuromorphic dynamics in a pure yet not obvious fashion and offer a direct dialogue with the machine and with richer soundscapes.  %It uses pose estimation to calculate a skeleton representation of the visitor. Though this is a classic approach, it is now being done with a neuromorphic camera and chip, generating very sparse data and consuming less energy. The installation has a dual purpose. It exposes the internals of the neuromorphic brain’s activity in a pure yet not obvious visual feedback. It also revisits the body controlled sound stylistic exercise with a more direct dialogue with the machine and with richer soundscapes.
\subsubsection{Le tourbillon de la vie}
This sound experience dives the visitor into a nostalgic atmosphere. Time is the binding factor between the visitor and the neuromorphic algorithm, which reacts on time sequences. The famous song from Jeanne Moreau is sung a capella, while sounds of pleasant remembrances pop up with the visitor’s arm and leg movements.
Movements to try out: left arm up, right arm up, both arms up, one foot away from the other, bend down.
\subsubsection{Tech now}
A closed feedback between the machine and the visitor creates a techno loop. Visitor’s arm movements trigger new tracks or generate effects, the level of activity raises percussions, The techno loop is built step by step in a more and more exciting sound.
Movements to try out: raise each arm, stretch arms horizontally, turn around, bend down.
\subsubsection{Screaching}
Here the connection between the visitor and the neuromorphic algorithm is direct. The visitor’s limb movements are transformed by the algorithm  into sounds, creating a unique dialogue.
Movements to try out: arm up, leg up, bend down.

\section{RESULTS}

\subsection{Pose estimation accuracy}

The four head outputs are combined to obtain predicted joint locations:
for each joint, heatmap values are masked to keep only pixels with values over a certain threshold $\tau=0.1$. The remaining pixels closest to the regression predictions are kept and multiplied by the model's down-sampling factor (4 in our experiments), to which we had the offset value, as mentioned in Sec.~\ref{subsection:architecture}.

We measure our accuracy with the dataset DHP19 \cite{Calabrese_2019_CVPR_Workshops} to compare to state of the art models. The chosen metric is the Mean Per Joint Prediction Error (MPJPE), computed as follows:
\begin{equation}
    MPJPE = \frac{1}{N}\Sigma_j^N|p_j-y_j|
\end{equation} 
with N=13 the number of joints and $|p_j-y_j|$ the distance in pixel between the predicted location of the joint $j$ and its actual location $y_j$. %Since we downsampled the input to 160x160, we multiply our results by a factor $260/160$ for state of the arts MPJPE comparison on the DHP19 resolution. 
 Tab.~\ref{table:accuracy} shows the comparison of our model with previous works. It outperforms the only other SNN implementation presented \cite{aydin2024hybridannsnnarchitecturelowpower}, which does not consider hardware constraints. The accuracy gap with state of the art model can be explained by the much smaller network size, reflected in the number of operations of Tab.~\ref{table:accuracy}.

\begin{figure}
     \centering
     \begin{subfigure}[b]{0.24\textwidth}
         \centering
         \includegraphics[width=\textwidth]{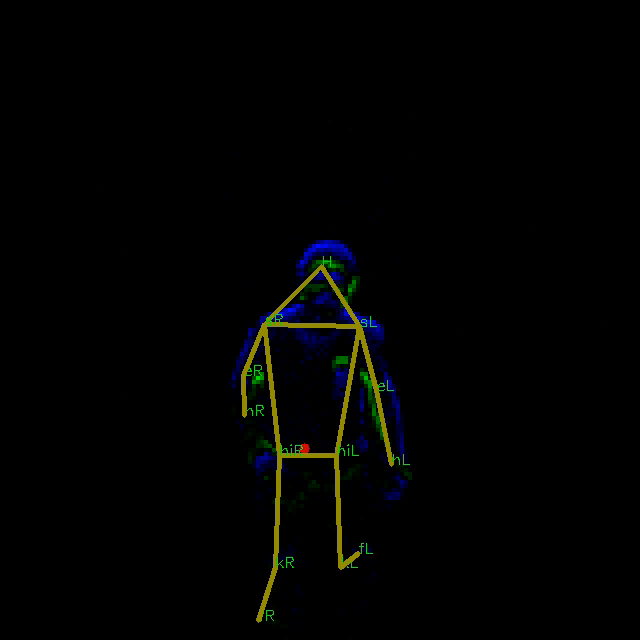}
         \subcaption{Missing joint}
     \end{subfigure}
     \hfill
     \begin{subfigure}[b]{0.24\textwidth}
         \centering
         \includegraphics[width=\textwidth]{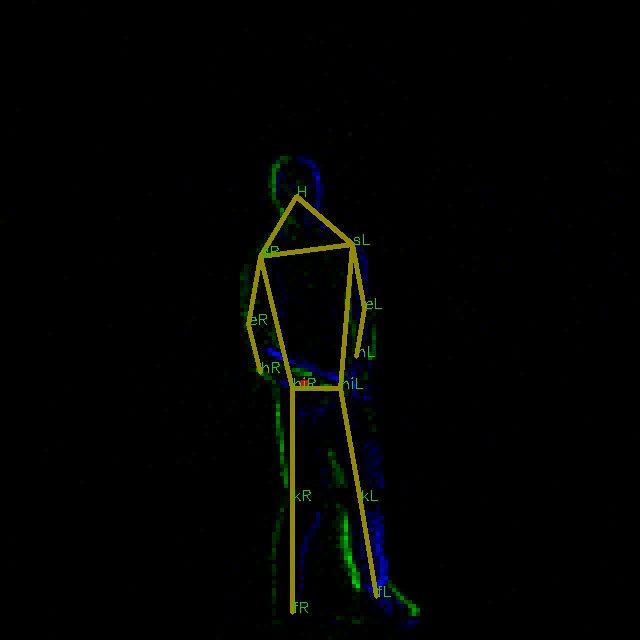}
         \subcaption{Average distribution}
     \end{subfigure}
     \begin{subfigure}[b]{0.24\textwidth}
         \centering
         \includegraphics[width=\textwidth]{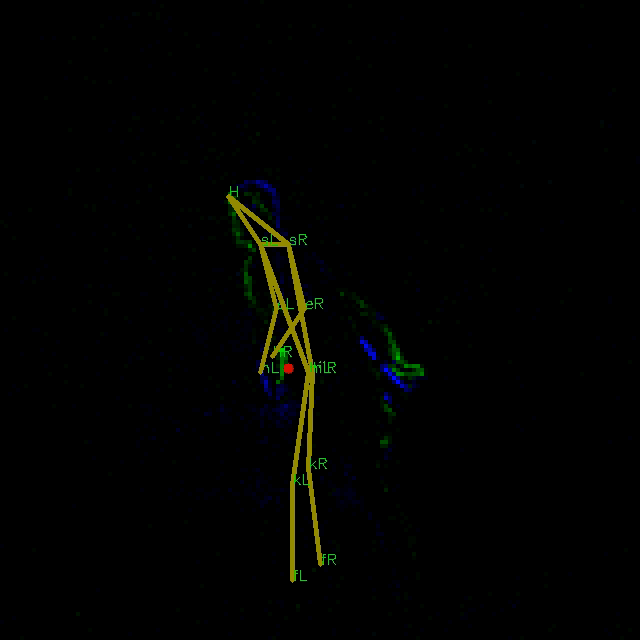}
         \subcaption{Body too rotated}
     \end{subfigure}
     \hfill
     \begin{subfigure}[b]{0.24\textwidth}
         \centering
         \includegraphics[width=\textwidth]{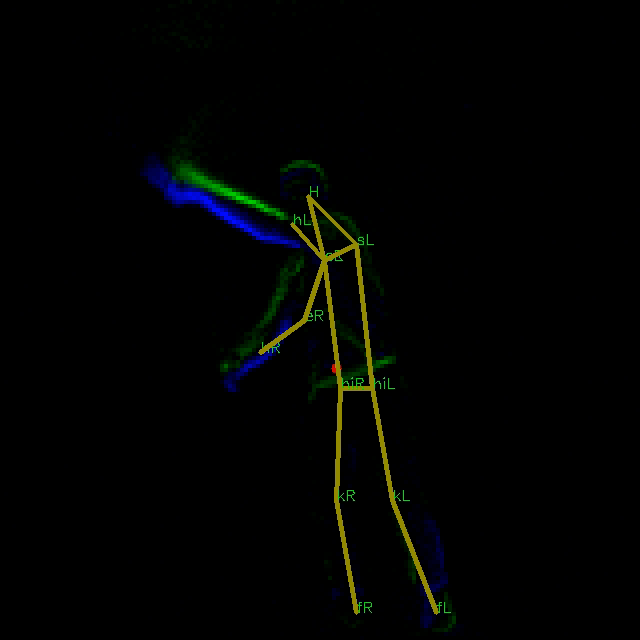}
         \subcaption{Limb too high}
     \end{subfigure}
        \caption{Four different HPE results overlayed on input events. Joint prediction locations are vertices of the skeleton and their name abbreviated (e.g. sL for shoulder Left). Lines drawn between connected joints are for visualisation only.}
        \label{fig:qualitative results}
\end{figure}

Qualitatively, the model has accurate predictions for small range movements close to the average distribution (a standing pose facing the camera). However, if the person does not move enough the event camera does not capture enough events on certain limbs, leading to unstable and incorrect predictions. Additionally, if the person's posture is too unusual, e.g. their legs are too high or the person is too rotated with respect to the camera, the model predicts closer positions to this previously described average position and confuses body's side. Fig.~\ref{fig:qualitative results} illustrates these results.
\begin{table}[htbp]
\centering
\caption{Results of Human pose estimation on DHP 19 and our new dataset, using MPJPE as accuracy metric (the lower the better) and the number of operations per second. The bottom section of the table shows model that can run on the Intel Loihi 2 chip}
\begin{tabular}{lccc}
\hline
Model & MPJPE & \#Operations (GFLOP/s)\\ 
\hline
Calabrese \cite{Calabrese_2019_CVPR_Workshops} & 7.67 & 255 MAC 0 AC\\
Baldwin \cite{TORE}& 5.62 & 949 MAC 0 AC\\
Goyal \cite{Goyal_2023_CVPR} & 6.28 & --\\
Ours - multihead (ANN) & 16.26 & 40 MAC 0 AC \\ %slurm-5714 and  utils/operations.py for operations computation
Aydin (Hybrid) \cite{aydin2024hybridannsnnarchitecturelowpower} & 5.08 & 233 MAC 79 AC\\
Aydin (SNN) \cite{aydin2024hybridannsnnarchitecturelowpower} & 20.27 & 0.5 MAC 121 AC\\
Aydin (ANN) \cite{aydin2024hybridannsnnarchitecturelowpower} & \textbf{4.85} & 2200 MAC 0 AC\\
\hline
Ours - heatmaps (SNN) &  13.73 & \textbf{16.7} MAC \textbf{0.7} AC\\ %slurm-4731.out for accuracy, slurm-4956.out for energy wiht torchprofile
Ours - multihead (SNN) & 12.07 & \underline{19.13} MAC \underline{0.7} AC\\ %slurm-4341.out for accuracy  %slurm-4956.out for energy with torchprofile\\ 
%Ours SNN - small multihead & 255 MAC 0 AC\\
%Ours SNN  - small & 255 MAC 0 AC\\
\hline
\end{tabular}
\label{table:accuracy}
\end{table}

\subsection{Energy and sparsity} \label{subsection:energy}
The network encoder is designed to run on the intel Loihi 2 chip and its decoder on a GPU.
The SNN encoder performs ACcumulate (ACs) operations, while the ANN decoder and heads are using Multiply and ACcumulates (MACs) operations. To estimate the average synaptic operations of a forward pass (with a number of time steps equals to the number of time bins per sample as explained in Sec.~\ref{subsection:sparse event based hpe}), we sum the number of operations over all layers. 
For a convolution layer, the number of operations $N_o$ can be written:
\begin{equation}
    N_o = \frac{\Sigma_{i=1}^N S_i}{N} * C_{out} * \frac{K^2}{s}
\end{equation}
where N is the size of the testing set and $S_i$ the input spike count of the sample $i$, $K$ is the kernel size, $s$ is the stride and $C_{out}$ the number of output channels. Similarly, for the decoder part, we have: 
\begin{equation}
    N_o = C_{in}*H_{out}*W_{out}  * C_{out} * K^2
\end{equation}
where $C_in$ is the number of output channels, and $H_{out}$ and $W_{out}$ the dimensions of the output feature map.
As shown in Tab.~\ref{table:accuracy}, the number of operations in our models is at least one order of magntude smaller than the state of the art. Comparing to theoretical MAC operations in an equivalent dense ANN gives us an average sparsity factor of 26.
 AC has a much lower energy consumption than MAC, that are widely operated in ANNs (for a 7nm CMOS processor, 0.38~$pJ$ instead of 1.69~$pJ$). Even if a proper power estimation would not only include operations cost, this difference added to the great sparsity obtained in the model still indicates that running models on neuromorphic hardware allows great energy savings.

\begin{figure}
     \centering
        \includegraphics[width=0.48\textwidth]{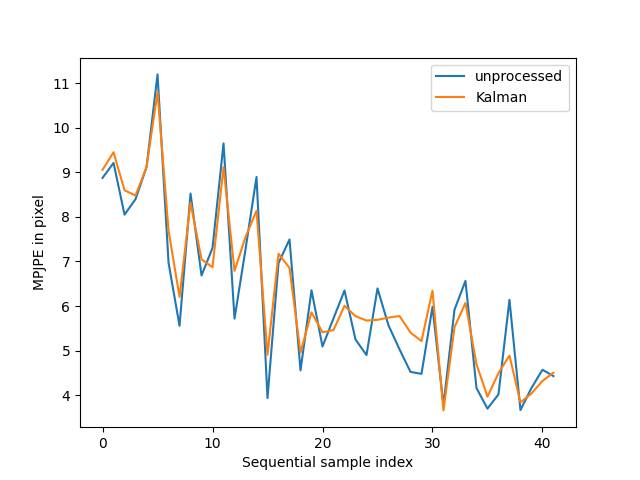}
        \caption{Impact of the Kalman filter (MPJPE over time).}
        \label{fig:Ablationkalman}
\end{figure}

\begin{table}[htbp] 
\centering
\caption{MPJPE and MACS on DHP19 for different time bins and neuron models. Only the spiking encoder is modified here.}
\label{table:binningimpact}
\begin{tabular}{lccc}
\hline
Model & time bins & MPJPE & sparsity\\ 
\hline
PLIF & 4 & 13.88& 31.33\\ %slurm-4921.out quantized not binary
PLIF & 8 & 13.21& 13.4\\ %slurm-4340.out quantized not binary
PLIF & 16 & \underline{12.44}& 19.78\\ %slurm-5739.out but epoch 16 quantized not binary
PLIF & 32 & \textbf{12.07}& 19.87\\ %slurm-4341.out and slurm-4933.out for eval quantized not binary
\hline
CUBA & 4&  NaN& NaN\\ %slurm-4840.out 
CUBA & 8&  13.55& 20.65\\ %slurm-4840.out binary
CUBA & 16&  13.75& \underline{35.79}\\ %slurm-4768.out binary
CUBA & 32& 14.24 & \textbf{39.39}\\ % binary %slurm-5655.out %slurm-4852.out
\hline
\end{tabular}
\end{table}

\subsection{Ablation}
To assess the key parameters of our model, we ran an ablation study on the following components:
\subsubsection{Kalman filter} we computed the MPJPE every 10ms for successive samples of one recording. We updated the Kalman states of each joint at the same frequency and measured the improved MPJPE. As shown in Fig.~\ref{fig:Ablationkalman}, after a few iterations, the error is consistently lower, with an average of 15.19 instead of 16.42 on our custom testing set. Visually, the Kalman filter has proven to give more stable and accurate predictions.
\subsubsection{Time bin resolution} we changed the time resolution of neuron dynamics and input events and compared the obtained accuracy and sparsity, as presented in Tab.~\ref{table:binningimpact}. The higher the number of time steps for one sample, the longer training and inferences take and the more operations may be necessary.For the PLIF implementations, MPJPE is improved with the number of time steps, however this is not satisfying considering the negative impact on latency. For the CUBA implementation, latency also increases with more time bins and the best accuracy is reached with only 8 time bins, suggesting that a finer time resolution is not beneficial. However, the network did not converge with a too coarse resolution of only 4 time bins.
\subsubsection{Artificial encoder} we trained an ANN version of our model as shown in Tab.~\ref{table:accuracy}. As expected, the ANN model presents much more operations and only uses MACS as explained in \ref{subsection:energy}. We also note that previous models beat our ANN version due to their much larger architecture. Surprisingly, the MPJPE is higher than using an SNN encoder.
\subsubsection{Neuron model} we evaluated the PLIF and CUBA neuron models described in Sec.~\ref{subsection:sparse event based hpe} as seen in Tab.~\ref{table:binningimpact}. We obtained more accurate results with the PLIF networks and a comparable sparsity.
\subsubsection{Multihead loss} as explained in Sec.~\ref{subsection:architecture}, we combined several heads to obtain our predictions, while previously cited models using event based data only rely on heatmaps. The multihead models were shown to be more accurate, as seen in Tab.~\ref{table:accuracy}. However, the network struggles to learn all losses at the first epoch. To start the training with an easier objective, we only use the joints and center heatmap losses for the first training epoch.
%\subsubsection{Architecture tuning} we tried different architectures to see how reducing the number of layers, thus reducing the energy consumption and the latency, impacts the accuracy of the model. As shown in \ref{table:accuracy}, the accuracy decreases in the smaller implementation.

\subsection{Artistic Exhibition}
Our artistic installation TONUS was showcased at the Festival der Zukunft in Munich on 27-30 June, 2024. This festival is a major event in Bavaria's scientific and artistic life. It is open to the general public (10000+ visitors) and presents deep tech innovations as well as their implications in society and art. Renowned media artists present their installations, influential entrepreneurs and scientists discuss future technology. Our installation was in the artistic area of the exhibition hall and raised a lot of interest from as varied profiles as tech journalists, tech companies, dancers, designers, researchers, students, families.
\begin{figure}
     \centering
     \begin{subfigure}[b]{0.45\textwidth}
         \centering
         \includegraphics[width=\textwidth]{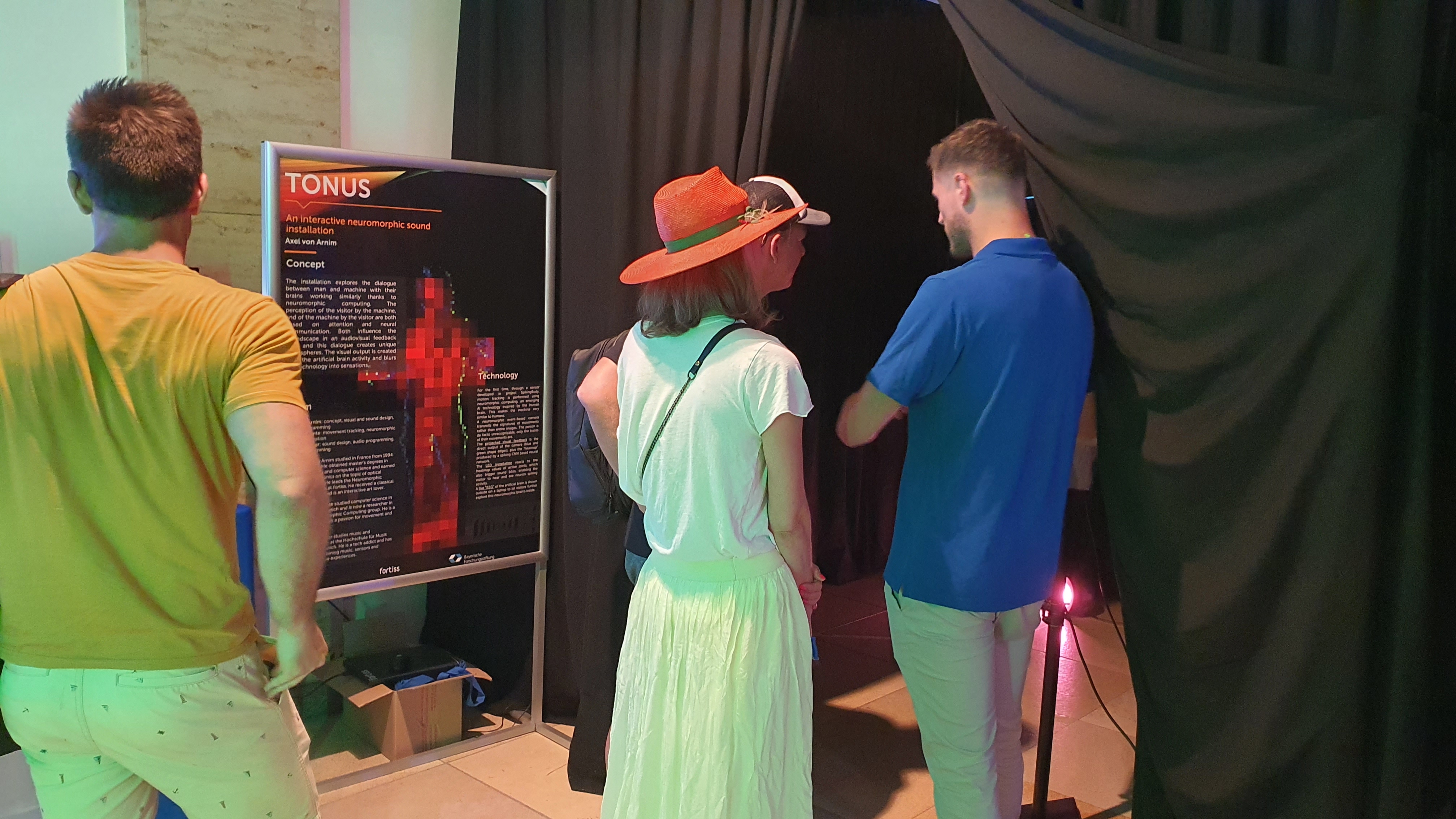}
     \end{subfigure}
     \hfill
     \begin{subfigure}[b]{0.45\textwidth}
         \centering
         \includegraphics[width=\textwidth]{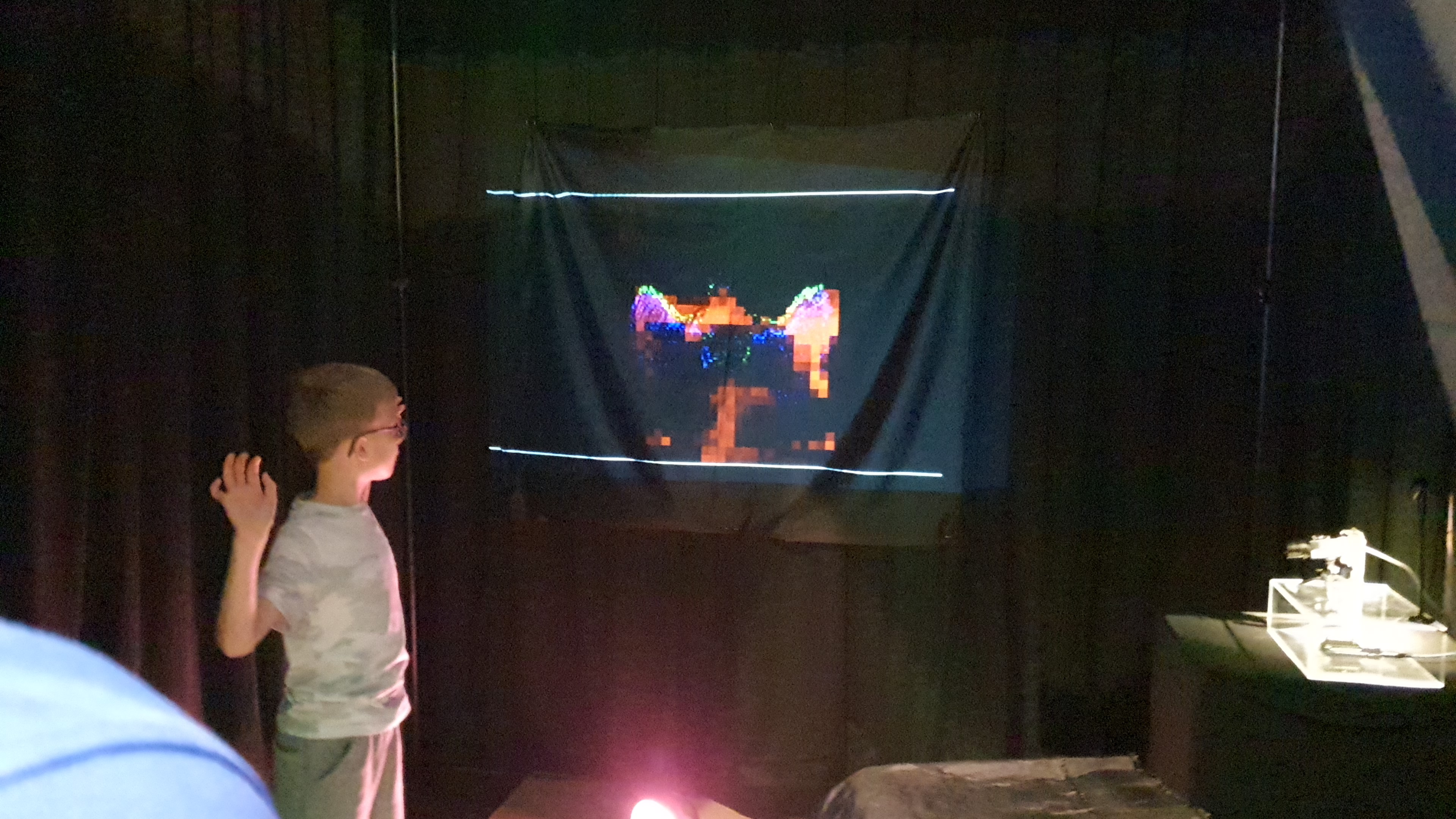}
     \end{subfigure}
        \caption{Visitors at our installation.}
        \label{fig:festival_photos}
\end{figure}
Fig.~\ref{fig:festival_photos} presents pictures of our installation but a video\footnote{\url{https://youtube.com/shorts/tkcpKS0wHVI}} gives a much more immersive impression with audio included.

\section{CONCLUSION AND DISCUSSION}

In the frame of this work, a SNN was designed and trained to extract features for a human pose estimation network, considering realistic neuromorphic hardware constraints. Though reaching smaller accuracies than state of the art networks, our network encoder is portable on neuromorphic hardware and has state of the art sparsity and operations count.
An artistic sound installation was exhibited, featuring the pose estimation sensor and establishing a link between sensory outputs and neuronal data processing. It aimed at giving the visitor an impression of biological brotherhood with the machine, its brain-like principles and intuitive interaction capabilities. Noteworthy, an art installation will be submitted to the 2025 IJCNN Exhibition in March, inspired by the prototype presented in this paper.
One key possible improvement is the real time port onto the neuromorphic chip Loihi 2, which was until now limited by the missing fast I/O interface. Furthermore, our current decoder is not currently runnable on the chip because of unsupported deconvolution layers. One could either implement corresponding synaptic connections manually or have a fully regression-based approach, taking inspiration from the current regression head of our model. This regression can be done with convolution or fully connected layers, already available on the chip. 
Network improvements could consider the temporal correlation between successive input samples, thus retaining information on the body pose for a longer time. This would potentially solve the non visible joint issue or provide a context for rare, out of distribution motions. \newline
Artistically, a more elaborated LED visual could provide a deeper understanding of the action/spike mapping and music-aware projections could be explored. Additionally, a camera could be added, enabling 3D positioning of the visitor, finer estimations and allow for richer interactions.

\section*{Author contributions}
Author 1 contributed the pose estimation algorithm, Author 2 contributed the artistic idea and design, Author 3 contributed to the pose estimation network architecture Author 2 contributed the audio engineering and artistic design.

\section*{Acknowledgment}
We warmly thank Dominik Giesriegl and Alex Vicar for their artistic and technical support in the early exploratory stages of this project. We also thank Marcella Toth for her work on the dataset and user interfacing. We thank Nils Kazinski, Soubarna Banik and Lars Benner for providing the new dataset and feedbacks.

\bibliographystyle{ieeetr}
\bibliography{bibliography.bib}

\end{document}